\documentclass[10pt,twocolumn,letterpaper]{article}
\usepackage{iccv}
\usepackage{times}
\usepackage{epsfig}
\usepackage{graphicx}
\usepackage{amsmath}
\usepackage{amssymb}

\usepackage{diagbox}
\usepackage{multirow}
\usepackage{booktabs}
\usepackage[table,xcdraw]{xcolor}
\usepackage[accsupp]{axessibility}
\usepackage[linesnumbered,vlined,ruled,commentsnumbered]{algorithm2e}

\usepackage[pagebackref=true,breaklinks=true,letterpaper=true,colorlinks,bookmarks=false]{hyperref}

\usepackage[capitalize]{cleveref}

\iccvfinalcopy 

\def\iccvPaperID{4842} 

\ificcvfinal\fi

\newcommand{\Amat}{{\boldsymbol A}}

\newcommand{\Kmat}[0]{{{\boldsymbol K}}}
\newcommand{\Lmat}[0]{{{\boldsymbol L}}}
\newcommand{\Mmat}[0]{{{\boldsymbol M}}}

\newcommand{\Qmat}[0]{{{\boldsymbol Q}}}

\newcommand{\Vmat}[0]{{{\boldsymbol V}}}
\newcommand{\Wmat}[0]{{{\boldsymbol W}}}
\newcommand{\Xmat}{{\boldsymbol X}}
\newcommand{\Ymat}[0]{{{\boldsymbol Y}}}
\newcommand{\Zmat}{{\boldsymbol Z}}

\newcommand{\xv}{\boldsymbol{x}}
\newcommand{\yv}{\boldsymbol{y}}

\newcommand{\zv}{\boldsymbol{z}}

\newcommand{\Deltamat}[0]{{\boldsymbol{\Delta}}}

\newcommand{\Sigmamat}{\boldsymbol{\Sigma}}

\newcommand{\Phimat}{\boldsymbol{\Phi}}

\newcommand{\Omegamat}{{\boldsymbol{\Omega}}}

\newcommand{\phiv}{\boldsymbol{\phi}}

\crefname{section}{Sec.}{Secs.}
\Crefname{section}{Section}{Sections}
\Crefname{table}{Table}{Tables}
\crefname{table}{Tab.}{Tabs.}
\newcommand*\tcircle[1]{%
  \raisebox{-0.001pt}{%
    \textcircled{\fontsize{6pt}{0}\fontfamily{phv}\selectfont #1}%
  }%
}

\begin{document}

\title{Deep Optics for Video Snapshot Compressive Imaging}
\vspace{-4mm}
\author{
Ping Wang$^{1,2}$ \qquad Lishun Wang$^{2}$ \qquad Xin Yuan$^{2,}$\thanks{Corresponding author.}\\
$^{1}$Zhejiang University \quad 
$^{2}$School of Engineering, Westlake University\\
{\tt\small 
\{wangping,wanglishun,xyuan\}@westlake.edu.cn}
}

\maketitle

\ificcvfinal\thispagestyle{empty}\fi

\begin{abstract}
Video snapshot compressive imaging (SCI) aims to capture a sequence of video frames with only a single shot of a 2D detector, whose backbones rest in optical modulation patterns (also known as masks) and a computational reconstruction algorithm. Advanced deep learning algorithms and mature hardware are putting video SCI into practical applications.
Yet, there are two clouds in the sunshine of SCI: $i)$ low dynamic range as a victim of high temporal multiplexing, and $ii)$ existing deep learning algorithms' degradation on real system.
To address these challenges, this paper presents a deep optics framework to jointly optimize masks and a reconstruction network.
Specifically, we first propose a new type of {\bf \em structural mask} to realize {\bf \em motion-aware} and {\bf \em full-dynamic-range} measurement. Considering the motion awareness property in measurement domain, we develop an {\bf \em efficient} network for video SCI reconstruction using Transformer to capture {\bf \em long-term temporal dependencies}, dubbed {\bf \em Res2former}. Moreover, {\bf \em sensor response} is introduced into the forward model of video SCI to guarantee end-to-end model training close to real system. 
Finally, we implement the learned structural masks on a digital micro-mirror device.
Experimental results on synthetic and real data validate the effectiveness of the proposed framework. We believe this is a milestone for real-world video SCI. The source code and data are available at \url{https://github.com/pwangcs/DeepOpticsSCI}.
\end{abstract}

\section{Introduction}
Capturing high-dynamic-range (HDR) and high-frame-rate (HFR) video is a long-term challenge in the field of computational photography. As an elegant solution of HFR, video snapshot compressive imaging (SCI) optically multiplexes a sequence of video frames, each of which is coded with a distinct modulation pattern (hereafter called mask), into a snapshot measurement of a two-dimensional (2D) detector, and computationally reconstructs a decent estimate of the original video from the measurement using an advanced algorithm.
In a nutshell, video SCI is a hardware-encoder-plus-software-decoder system and its performance mainly depends on mask and reconstruction algorithm.

\begin{figure}[t]
   \centering
    \includegraphics[width=0.8\linewidth]{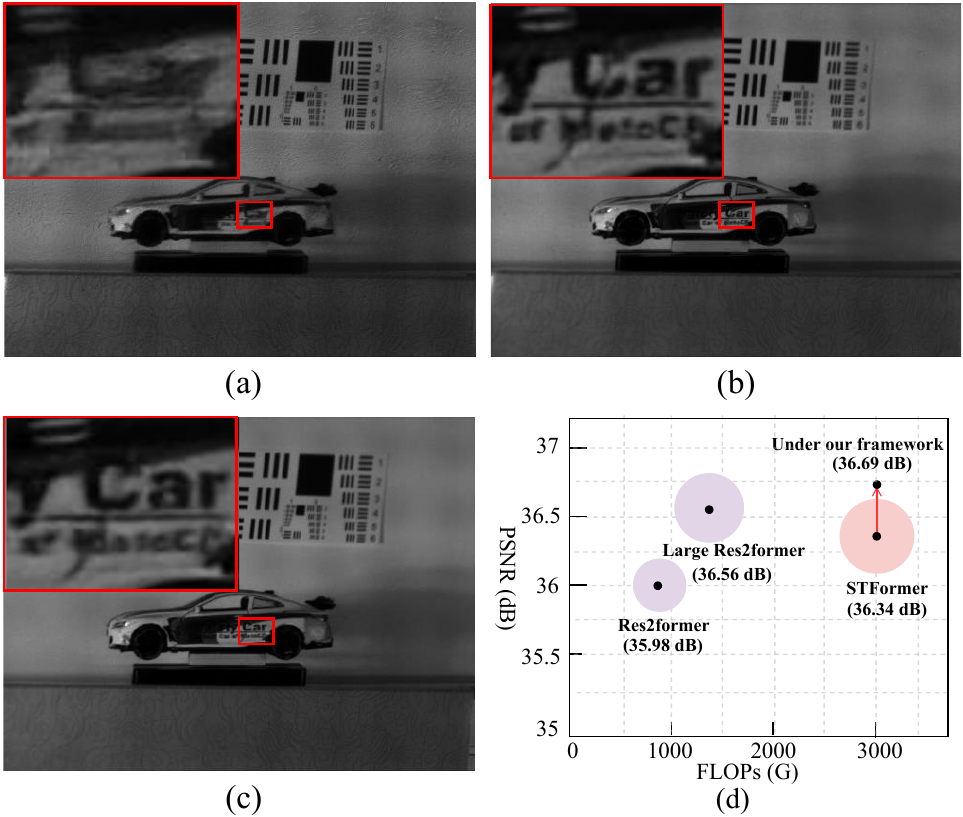}
    \vspace{-1mm}
    \caption{The proposed deep optics framework brings a significant improvement for real-world video SCI as demonstrated in real results (a), (b), and (c), got by previous SOTA STFormer~\cite{wang2022spatial}, current STFormer under our framework, and our Res2former, respectively. (d) summarizes the comparison between Res2former and STFormer in terms of PSNR (vertical axis), {FLOPs} (horizontal axis), and Parameters (circle radius). The proposed Res2former achieves competitive performance ($35.98$ dB) with only $28.15\%$ FLOPs and $56.57\%$ parameters of STFormer ($36.34$ dB). By increasing parameters to STFormer's level, large Res2former can lead to a better performance ($36.56$ dB). By the way, STFormer under our framework can increase by $0.35$ dB.}
 \label{fig:ForwardModel}
 \vspace{-5mm}
 \end{figure}

For the hardware encoder, random binary mask has been widely used in both simulation and real video SCI systems, often implemented in a digital micro-mirror device (DMD)~\cite{Qiao2020_APLP,Qiao2020_CACTI} or liquid crystal on silicon (LCOS)~\cite{Reddy11_CVPR_P2C2,Hitomi11_ICCV_videoCS,liu2013efficient}.
Recently, learned binary mask was also implemented in programmable pixel sensors~\cite{martel2020neural}. For the software decoder, 
it is an ill-posed inverse problem to retrieve high-fidelity video from the captured single measurement and various reconstruction methods~\cite{Yuan16ICIP_GAP,Liu19_PAMI_DeSCI,Ma19ICCV,Qiao2020_APLP,Cheng20ECCV_Birnat,Wang2021_CVPR_MetaSCI,Cheng2021_CVPR_ReverSCI,wu2021dense,wang2022spatial,meng2023deep} have been developed to solve it in recent years. 
Conventional optimization algorithms adopt hand-crafted priors, \eg, total variation~\cite{Yuan16ICIP_GAP} and non-local self-similarity~\cite{Liu19_PAMI_DeSCI}, to confine the solution to the desired signal space. But optimization-based methods commonly require a long running time to get usable results.
With the powerful generalization ability of deep neural networks (DNNs), deep learning methods have been increasingly developed and achieved excellent results in a little inference time, usually designed as an end-to-end (E2E) network, \eg, E2E-CNN~\cite{Qiao2020_APLP}, BIRNAT~\cite{Cheng20ECCV_Birnat}, MetaSCI~\cite{Wang2021_CVPR_MetaSCI}, RevSCI~\cite{Cheng2021_CVPR_ReverSCI}, STFormer~\cite{wang2022spatial}, or a deep unfolding network, \eg, GAP-net~\cite{meng2023deep}, ADMM-Net~\cite{Ma19ICCV}, SCI3D~\cite{wu2021dense}, ELP-Unfolding~\cite{yang2022ensemble}.
Despite these remarkable advances, particularly in deep reconstruction algorithms, there are still some practical challenges in putting video SCI into our daily life.

{\em Due to the limited bit depth of image sensors, the higher temporal multiplexing, the lower dynamic range}. For an video SCI camera using random binary masks, the measurable brightness values of video frames is approximately equal to ${{{2^{\kappa  + 1}}} \mathord{\left/ {\vphantom {{{2^{\kappa  + 1}}} B}} \right. \kern-\nulldelimiterspace} B}$, far less than the available brightness values of image sensor $ {2^\kappa }$, where $B$ (usually $8 \!\le B \!\le 50$) and $\kappa$ denote compressed frames and sensor bit depth, respectively. We take $8$-frame video SCI camera equipped with a typical $8$-bit-depth image sensor as an example, namely, $8$ video frames are compressed into a single image with $256$ available brightness values during measurement. If using random binary masks that take values of `1' or `0' with equal probability, at each spatial position, half of $8$ frames are integrated into one pixel along temporal dimension with high probability. In this case, each frame can only be represented by $64$ brightness values, which is calculated by ${{256} \mathord{\left/ {\vphantom {{256} 4}} \right. \kern-\nulldelimiterspace} 4} \!=\! 64$.
Obviously, there is a significant gap between the wide range of brightness variations in natural scenes and the very limited dynamic range in previous video SCI. Such a practical problem is also widely rooted in other compressive imaging systems, \eg, spectral SCI~\cite{Gehm07_DDCASSI}, compressive light field imaging~\cite{marwah2013compressive}, and single-pixel imaging~\cite{duarte2008single}.

{\em Without considering sensor response, existing deep reconstruction networks have a great performance degradation when used in real system}. As is well known, the performance of DNNs is closely related to the used training dataset. Without available specialized datasets, the forward model of video SCI usually need to be mathematically formulated to synthesize the training dataset from a public HFR video dataset. Accordingly, deep reconstruction networks have a high dependence on the forward model. Unfortunately, previous forward model only considers {\em optical transmission and modulation} but overlooks {\em sensor response} characterizing the used image sensor, meaning that there is a gap between previous forward model and real system. As a result, previous deep reconstruction networks show excellent performance on synthetic data but degraded performance on real data.

To address the above challenges, a deep optics framework is proposed to improve the performance of real-world video SCI. The contributions of this work are summarized as follows.
\begin{itemize}
\setlength{\itemsep}{0pt}
   \item Unlike widely-used random binary mask, a new type of {\em structural mask} is presented to realize {\em motion-aware} and {\em full-dynamic-range} (FDR) measurement. Motion-aware measurement contributes to video SCI reconstruction. To our best knowledge, we are the first to enable FDR video SCI.
   \item Considering the motion-aware property in the encoder, we tailor an {\em efficient} reconstruction network, dubbed Res2former, as the video SCI decoder by using Transformer to capture {\em long-term temporal dependencies}. Compared with the state-of-the-art (SOTA) network STFormer~\cite{wang2022spatial}, Res2former is highly lightweight but provides competitive performance. 
   \item We propose a deep optics framework to jointly optimize the proposed structural mask and reconstruction network, in which {\em sensor response} is introduced to guarantee end-to-end (E2E) training close to real system. Under this framework, Res2former and previous reconstruction networks achieve significant improvement on synthetic data and real data.
\end{itemize}

\section{Related Work}

\noindent{\bf{Deep optics}.}
Deep optics takes the idea of jointly optimizing optics and algorithm to improve various computational imaging systems, \eg. microscopy~\cite{nehme2020deepstorm3d}, HDR imaging~\cite{martel2020neural,sun2020learning,metzler2020deep}, depth imaging~\cite{yoshida2018joint,chang2019deep,wu2019phasecam3d}, single-pixel imaging~\cite{horisaki2020deeply}, light field imaging~\cite{inagaki2018learning}, and compressing imaging~\cite{ILIADIS2020102591,yoshida2018joint,martel2020neural,Li2020ICCP,Zhang22Optica,vargas2021time}.
Mask optimization for video SCI has been increasingly studied under hardware constraints~\cite{yoshida2018joint,martel2020neural,Li2020ICCP}. 
Based on an emerging programmable sensor SCAMP-5, a hand-held video SCI camera~\cite{martel2020neural} has recently developed but its spatial-temporal resolution is very limited.
These works attached great importance to the implementation of binary mask by using some heuristic sensors. This paper aims to the performance of real-world video SCI. To our best knowledge, we are the first to optimize more challenging structural mask and remove the incompatibility between temporal multiplexing and dynamic range.

\noindent{\bf{Video SCI reconstruction}.}
Video SCI reconstruction algorithms can be classified into regularization-based methods and learning-based methods. Regularization-based methods combine the idea of iterative optimization, \eg, generalized alternating projection (GAP)~\cite{liao2014generalized} or alternating direction method of multipliers (ADMM)~\cite{Boyd11ADMM}, with various prior knowledge, \eg, total variation (TV)~\cite{Yuan16ICIP_GAP} and non-local low rank~\cite{Liu19_PAMI_DeSCI}. They provide usable results in an unsupervised manner but cannot balance fidelity and speed. In recent years, kinds of learning-based methods have been developed for high fidelity and low inference time. Recently, an E2E network STFormer~\cite{wang2022spatial} has achieved the state-of-the-art (SOTA) results using temporal and spatial Transformer, but at the cost of high parameters and complexity. In addition to E2E networks, \eg, E2E-CNN~\cite{Qiao2020_APLP}, BIRNAT~\cite{Cheng20ECCV_Birnat}, MetaSCI~\cite{Wang2021_CVPR_MetaSCI}, RevSCI~\cite{Cheng2021_CVPR_ReverSCI}, deep unfolding networks, \eg, GAP-net~\cite{meng2023deep}, ADMM-Net~\cite{Ma19ICCV}, SCI3D~\cite{wu2021dense}, ELP-Unfolding~\cite{yang2022ensemble}, and plug-and-play (PnP) algorithms, \eg, PnP-FFDNet~\cite{9156491} and PnP-FastDVDnet~\cite{9495194}, have been developed by combining an iterative optimization framework with convolutional neural networks or a deep image denoiser. Both regularization-based methods and learning-based methods aim to solve the ill-posed inverse problem of video SCI forward model, thus their performance is susceptible to this model. Previous forward model only considers optical transmission and modulation but overlooks sensor response in practice. As a result, existing reconstruction networks lead to excellent results in synthetic data rather than real data.

\section{Video SCI: from Theory to Practice}
Aiming to move one step further towards real-world video SCI, we hereby make a wide appeal for modeling video SCI under hardware constraints and employing structural mask instead of random binary mask.
\begin{figure}[tbp!]
\centering
\includegraphics[width=1\linewidth]{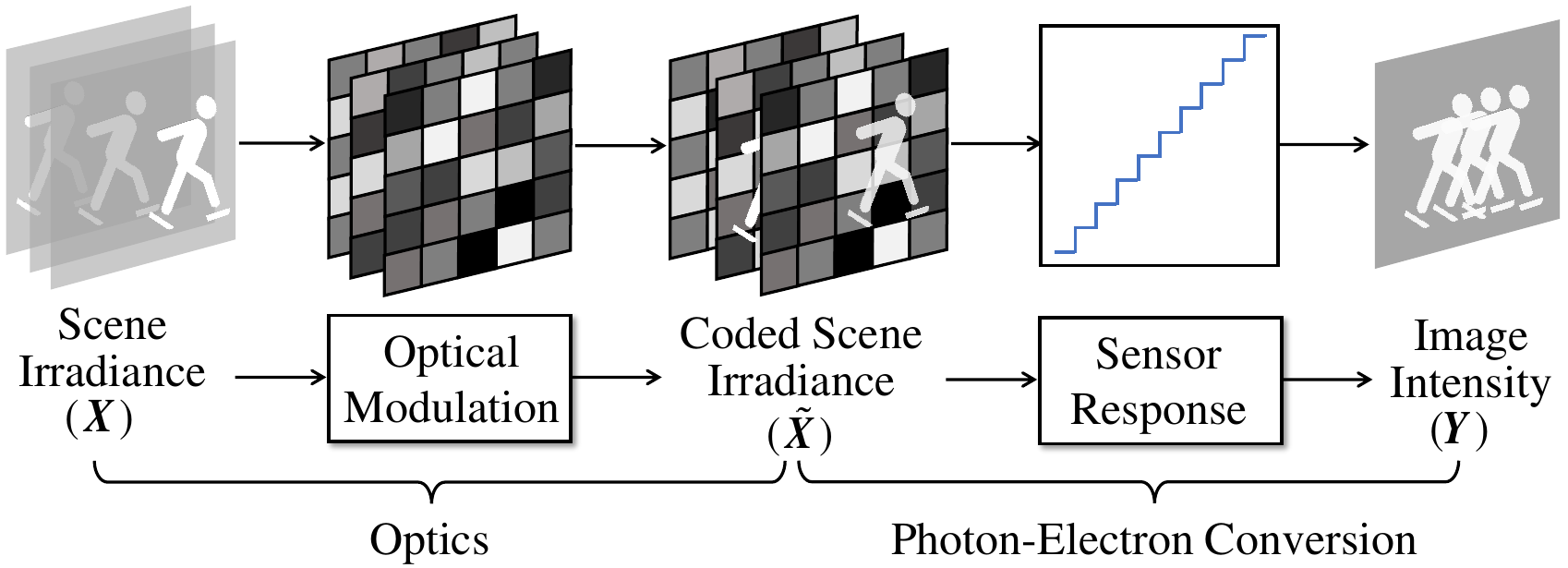}
\vspace{-3mm}
\caption{Illustration of video SCI encoder. High-speed scene is first optically modulated with temporally-varying masks and then integrated into a single digital image (\ie, snapshot measurement) through an off-the-shelf image sensor. In the process, optical modulation and sensor response are two key ingredients.}
\vspace{-5mm}
\label{fig:forward}
\end{figure}
\subsection{Mathematical Model of Practical Video SCI}
As shown Fig.~\ref{fig:forward}, video SCI encoder is mainly composed of optical modulation and sensor response. In the video SCI decoder, a reconstruction algorithm is employed.

{\noindent \bf Optical modulation}. By implementing temporally-varying masks ${\Mmat}(u,v,t)$ on $B$ discrete time slots ($1\!\le\! t \!\le\! B$), a dynamic scene irradiance $\Xmat(u,v,t)$ is modulated into the coded spatial-temporal irradiance ${\tilde \Xmat}(u,v,t)$ by
\begin{equation}\label{eq:modulation}
\setlength{\abovedisplayskip}{6pt}
\setlength{\belowdisplayskip}{6pt}
{\tilde \Xmat}(u,v,t) =  {\Mmat}(u,v,t) \odot \Xmat(u,v,t),
\end{equation}
where $(u,v,t)$ denotes the spatial-temporal coordinate and $\odot$ denotes the Hadamard (element-wise) product.

{\noindent \bf Sensor response}. Given an image sensor, ${\tilde \Xmat}(u,v,t)$ is integrated as a single digital image $\Ymat(u,v)$ (\ie, snapshot measurement) by
\begin{equation}\label{eq:response}
\setlength{\abovedisplayskip}{6pt}
\setlength{\belowdisplayskip}{6pt}
\Ymat(u,v)= {\cal R} \left[ {\sum_{t=1}^B{{\tilde \Xmat}(u,v,t)}}\right] + \Zmat(u,v), 
\end{equation}
where ${\cal R}$ represents the mapping function of from scene irradiance to image pixels and $\Zmat(u,v)$ denotes the noise originated from measurement, read-out, \etc. Defining the vectorization operation on a matrix as ${\tt vec}(\cdot)$, we can rewrite Eq.~\eqref{eq:response} into the following vectorized form:
\begin{subequations}
\setlength{\abovedisplayskip}{5pt}
\setlength{\belowdisplayskip}{5pt}
   \begin{align}
      &\yv = {\cal R} \circ {\cal H}(\xv) + \zv, \label{eq:forward2} \\
      &s.t. \quad {\cal H}(\xv) = {\Phimat \xv} , \label{eq:forward1}
   \end{align}
\label{eq:forward}
\end{subequations}
where $\xv \!=\! {\tt vec}(\Xmat)$, $\yv \!=\! {\tt vec}(\Ymat)$, $\zv \!=\! {\tt vec}(\Zmat)$, and $\Phimat \!=\! \left[ {\tt diag}\left({\tt vec}({\Mmat}(:,:,1)) \right), \cdots, {\tt diag}\left({\tt vec}({\Mmat}(:,:,B)) \right) \right]$. 

{\noindent \bf Computational reconstruction}.
Provided with the used $\Phimat$, a regularization-based or learning-based reconstruction algorithm ${\cal D}$ is employed to retrieve a decent estimate ${\hat \xv}$ of $\xv$ from $\yv$ by
\begin{equation}\label{eq:decoder}
\setlength{\abovedisplayskip}{5pt}
\setlength{\belowdisplayskip}{5pt}
{\hat \xv} = {\cal D} (\yv) = {\cal D} \circ {\cal R} \circ {\cal H}(\xv).
\end{equation}
In general, ${\cal R}$ can be modeled as a combination of non-linear response function $f$, out-of-range clipping function $g$, and quantization function $h$, \ie, ${\cal R} \!=\! h \circ g \circ f $, leading to non-linearity, saturation error, and quantization error, respectively. These functions are generally inevitable to transform real-value scene irradiance into digital image brightness. In most industrial cameras, the non-linearity function $f$ can be corrected to be linear, thus ${\cal R}$ is simplified as ${\cal R} = h \circ g $ in this paper.

Previous works~\cite{Yuan16ICIP_GAP,Liu19_PAMI_DeSCI,Ma19ICCV,Qiao2020_APLP,Cheng20ECCV_Birnat,Wang2021_CVPR_MetaSCI,Cheng2021_CVPR_ReverSCI,wu2021dense,wang2022spatial,meng2023deep} view Eq.~\eqref{eq:forward1}, only considering optical modulation and sensor integration, as the forward model of video SCI. By introducing the complete sensor response, we present the forward model in Eq.~\eqref{eq:forward} closer to real system.
\begin{figure}[t]
   \centering
   \vspace{-3mm}
   \includegraphics[width=0.8\linewidth]{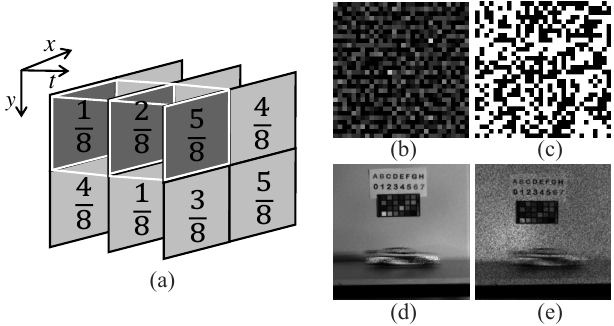}
   \vspace{-1mm}
   \caption{Proposed structural mask (b) \vs widely-used random binary mask (c). As demonstrated in (a), structural mask values represent the transmittance of incident light and the sum of values across temporal dimension is $1$. It lead to the motion-aware measurement (d), containing more visual information than the measurement (e) captured by random binary mask.}
   \vspace{-3mm}
\label{fig:mask}
\end{figure}

\subsection{Proposed Structural Mask}
As mentioned previously, there is an incompatibility between temporal multiplexing and dynamic range in existing works due to the use of random binary mask. Here, we propose a new type of structural mask to realize full-dynamic-range (FDR) and motion-aware measurement for video SCI, which is mathematically defined as
\begin{subequations}\label{eq:mask}
{
\setlength{\abovedisplayskip}{5pt}
\setlength{\belowdisplayskip}{-3pt}
\begin{equation}
\label{eq: mask1}
{{\mathcal{\Mmat}}_{\lambda} (u,v,t)} \!\in\! \left\{ {\begin{array}{*{20}{c}}
{\{ 0,{1 \mathord{\left/
 {\vphantom {1 {{2^\lambda }}}} \right.
 \kern-\nulldelimiterspace} {{2^\lambda }}},...,1 \!-\! {1 \mathord{\left/
 {\vphantom {1 {{2^\lambda }}}} \right.
 \kern-\nulldelimiterspace} {{2^\lambda }}}\} \quad \lambda  \!\ge\! 2} \\ {\{ 0,1\}  \quad \quad \quad \quad \quad \quad \quad\quad \lambda  \!=\! 1} \end{array}} \right.
\end{equation}
}
{
\setlength{\abovedisplayskip}{-3pt}
\setlength{\belowdisplayskip}{5pt}
\begin{equation}
\label{eq: mask2}
s.t. \quad \sum_{t=1}^B {\Mmat}_{\lambda}(:,:,t) = {\bf 1},
\end{equation}
}
\end{subequations}
\noindent where ${\lambda}$ denotes the bit depth of mask.
Unlike widely-used binary mask, the proposed mask has two attributes: {\em discretization} and {\em structuralization}.
In Eq.~\eqref{eq: mask1}, discretization indicates that the mask can only take binary ($\lambda\!=\!1$) or grayscale ($\lambda \!\ge\! 2$) values. In Eq.~\eqref{eq: mask2}, structuralization indicates that, at all spatial positions, the sum across temporal dimension is fixed to $1$, also demonstrated in Fig.~\ref{fig:mask} (a).
Due to the undesirable performance of $1$-bit (\ie, binary) structural mask (see Tab.~\ref{tab:ablation}), we mainly focus on the setting of $\lambda \!\ge\! 2$ in this paper.
Structural mask can also be easily implemented in an off-the-shelf spatial light modulator (\eg, DMD) at the cost of decreasing the pattern refresh rate. Fortunately, current DMDs' pattern refresh rate is high enough for video SCI. Taking DLP7000 DMD~\footnote{\url{https://ti.com/product/DLP7000}}
as an example, the maximal pattern refresh rate is $32552$ or $4069$ for $1$-bit mask or $8$-bit mask, respectively.

{\noindent \bf Full dynamic range (FDR)}.
The proposed structural mask is capable of removing the incompatibility between temporal multiplexing and dynamic range, rooted in previous video SCI using random binary mask.
Taking an $8$-bit image sensor as an example, it can record scene irradiance within brightness range $[0, 1, \dots, 255]$. 
For $8$-frame video SCI (\ie, $B\!=\!8$), the sum of random binary mask across temporal dimension is approximate to $4$, equivalent to that almost $4$ video frames are integrated into a single image within brightness range $[0, 1, \dots, 255]$. Accordingly, the brightness range of each video frame is limited in $[0, 1, \dots, 63]$, leading to a low dynamic range. Clearly, it cannot meet the wide range of brightness variations in natural scenes and worsen along with larger $B$. Using the proposed structural mask, each pixel of captured measurement is the weighted pixel sum of video frames across temporal dimension and the total weight is $1$. It means that the brightness range of measurement is equal to that of each video frame regardless of $B$. 
Therefore, the proposed structural mask keeps the dynamic range of video SCI in line with that of the used image sensor, meaning full dynamic range (FDR). 

{\noindent \bf Motion-aware measurement}. As shown in Fig.~\ref{fig:mask} (d), the motionless objects, background, and motion trajectory could be greatly recorded in the measurement captured by structural mask. We refer to it as motion-aware measurement.
Such measurement can be viewed as a coarse estimate of original video frames. Generating the network input from a coarse estimate is essential in nearly all impressive video SCI reconstruction networks~\cite{Cheng20ECCV_Birnat,Wang2021_CVPR_MetaSCI,Cheng2021_CVPR_ReverSCI,wu2021dense,yang2022ensemble,wang2022spatial}. Unlike our direct acquisition by optics, previous works get the coarse estimate by idealizing video SCI forward model as Eq.~\eqref{eq:forward1} and then computing
$(\Phimat {\Phimat^{\top}})^{-1}{\cal H}(\xv)$.
But their estimate becomes $(\Phimat {\Phimat^{\top}})^{-1}\yv$ in practice.
The mismatch in input initialization also makes for previous network's performance degradation in real system.

\section{Deep Optics Framework for Video SCI}
\begin{figure*}[th]
   \centering
    \vspace{0mm}
    \includegraphics[width=1\linewidth]{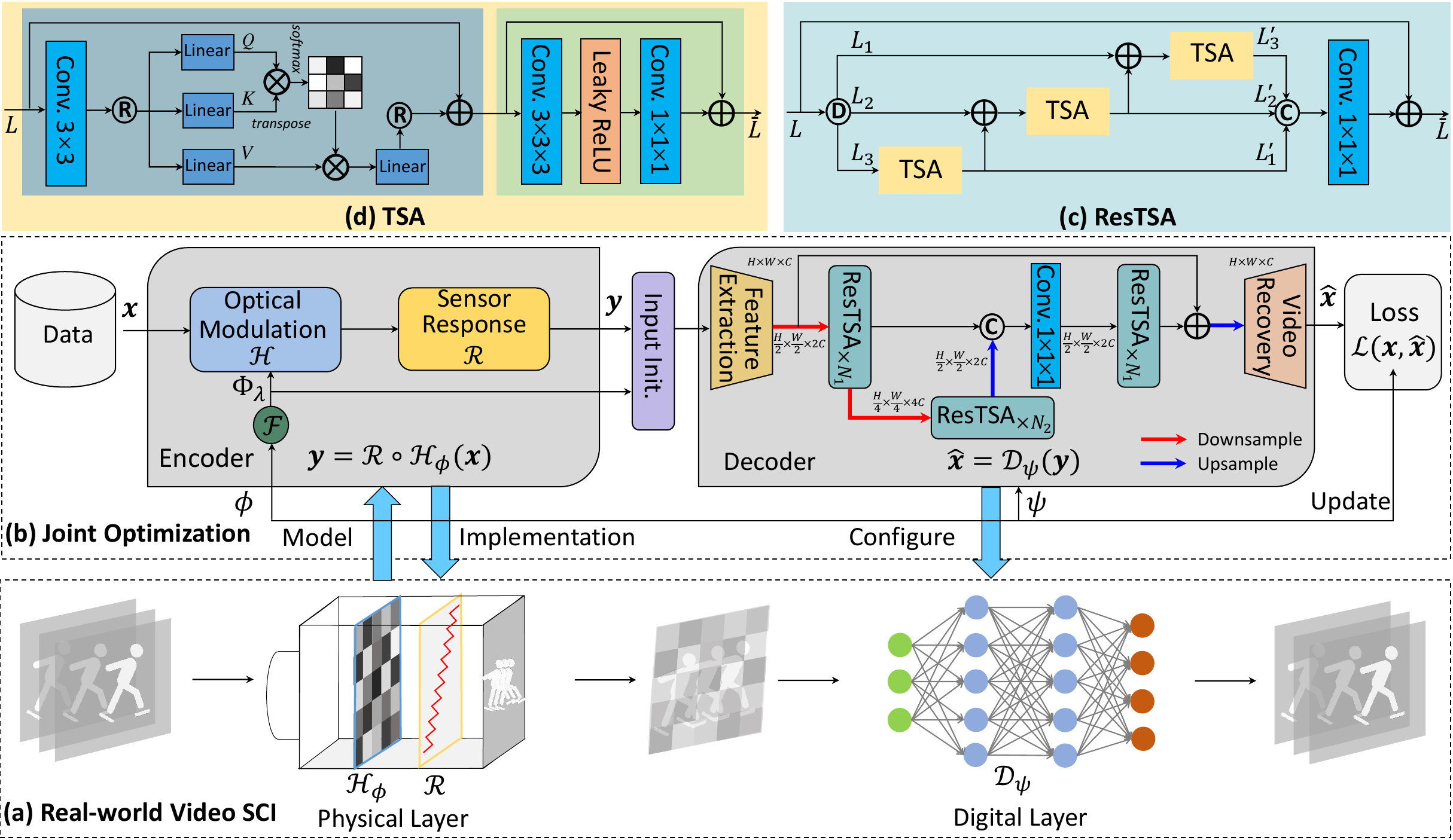}
    \vspace{-3mm}
    \caption{Deep optics framework for the joint optimization of structural mask and a deep reconstruction network. $\oplus$, $\tcircle{C}$/$\tcircle{D}$, and $\otimes$ denote element-wise addition, channel concentration/division, and matrix multiplication, respectively. For clarity, ResTSA module is depicted as a $3$-branch structure. In the decoder, the number of ResTSA modules can be adjusted. By default, $(N_1, N_2) \!=\! (3,3).$}
    \vspace{-4mm}
 \label{fig:framework}
 \end{figure*}
Previous video SCI reconstruction networks~\cite{Qiao2020_APLP,Cheng20ECCV_Birnat,Wang2021_CVPR_MetaSCI,Cheng2021_CVPR_ReverSCI,wang2022spatial,meng2023deep,Ma19ICCV,wu2021dense,yang2022ensemble} were trained on the impractical forward model in Eq.~\eqref{eq:forward1} and thus achieved impressive performance in simulation rather than real system. To bridge the performance gap, we propose an E2E deep optics framework to jointly optimize structural mask and a reconstruction network under hardware constraints.
\subsection{Overall Architecture}
As shown in Fig.~\ref{fig:framework} (a), a real-world video SCI system is composed of a temporal multiplexing camera for encoding in the physical layer and a regularization-based or learning-based reconstruction algorithm for decoding in the digital layer.
Due to the lack of specialized video SCI dataset, the forward model in Eq.~\eqref{eq:forward} is proposed to mimic the encoding camera that synthesizes measurement-video ($\yv$, $\xv$) pairs as training dataset. Sensor response is introduced to be closer to real camera.

As shown in Fig.~\ref{fig:framework} (b), the joint optimization framework includes a modeled encoder and a designed deep decoder.
The structural mask in the encoder and a deep reconstruction network as the decoder are jointly optimized by the following loss function:
\begin{equation}
\label{eq: loss}
\setlength{\abovedisplayskip}{5pt}
\setlength{\belowdisplayskip}{5pt}
   \mathop {\arg \min }\limits_{\left\{ {\phi ,\psi} \right\}} {\sum_{k = 1}^K {\left\| {{{\cal D}_\psi } \circ {\cal R} \circ {{\cal H}_\phi }({\xv_k}) - {\xv_k}} \right\|_2^2}},
\end{equation}
where $K$ denotes the number of training samples, $\phi$ and $\psi$ represent the parameters of structural mask and deep decoder ${\cal D}$, respectively.
The sensor response ${\cal R}$ is modeled as 
${\cal R}(x) \!=\! {{\left\lfloor {{255} \cdot x \!+\! 0.5} \right\rfloor } \mathord{\left/
{\vphantom {{\left\lfloor {{255} \cdot x + 0.5} \right\rfloor } {{2^d }}}} \right.
\kern-\nulldelimiterspace} {{255}}}$. As a hard thresholding function, ${\cal R}$ doesn't yield useful gradients and it follows the training strategy of mask optimization.
The proposed structural mask is updated along with $\phiv$ and the deep decoder ${\cal D}$ is a two-level U-shaped network with temporal self-attention mechanism, which are detailedly introduced in Sec.~\ref{sec:mask} and Sec.~\ref{sec:decoder}, respectively. 

\subsection{Structural Mask Optimization}
\label{sec:mask}
{
\begin{algorithm}[ht]
\caption{Structural Mask Training}
\label{alg:mask}
\SetKwInput{KwInput}{Input}                
\SetKwInput{KwOutput}{Output}              
\DontPrintSemicolon

\KwInput{A learnable mask ${{\Mmat}'} \!\in\! [0,1]$ with a size of $B \!\times\! H\!\times\! W$ and the desired bit depth $\lambda \!\ge\! 2$.}
\KwOutput{A $\lambda$-bit structural mask ${{\Mmat}_\lambda}$. }

\SetKwFunction{Forward}{$\mathcal{F}$}
\SetKwFunction{Backward}{$\mathcal{G}$}
\SetKwProg{Fn}{Forward}{:}{}
\Fn{\Forward{${{\Mmat}'}$}}
{
\tcc{Discretization}
$L \longleftarrow {2^{\lambda}}$\;
${{\Mmat}} \longleftarrow {{\left\lfloor {\Mmat'  \cdot L + {\boldsymbol {0.5}}} \right\rfloor } \mathord{\left/{\vphantom {{\left\lfloor {\Phi  \cdot L + {\boldsymbol {0.5}}} \right\rfloor } L}} \right.\kern-\nulldelimiterspace} L}$\;
${{\Mmat}}[{{\Mmat}} == 1] \longleftarrow 1 - {1 \mathord{\left/{\vphantom {1 L}} \right.\kern-\nulldelimiterspace} L}$ \;
${{\Mmat}}[:,\mathtt{sum}({{\Mmat}},0) == 0] \longleftarrow {1 \mathord{\left/{\vphantom {1 L}} \right.\kern-\nulldelimiterspace} L}$\;
\tcc{Structuralization}
$\Omegamat \longleftarrow \mathtt{sum}({{\Mmat}},0)$\;
$\Sigmamat \longleftarrow {L} \cdot (\Omegamat - {{\boldsymbol 1}})$\;
\For(){$0 \le k < B$}
{
$\Wmat \longleftarrow {{\Mmat}[k] \oslash {\Omegamat}}$\;
$\Deltamat  \longleftarrow  {{\left\lfloor \Sigmamat \odot \Wmat + {\boldsymbol 0.5} \right\rfloor} \mathord{\left/{\vphantom {{{\left\lfloor \Sigmamat \odot \Wmat + {\boldsymbol 0.5} \right\rfloor}} {L}}} \right.\kern-\nulldelimiterspace} {L}}$\;
${{\Mmat}_\lambda}[k] \longleftarrow {{\Mmat}}[k] - \Deltamat$\;
$\Omegamat \longleftarrow \Omegamat -{{\Mmat}}[k]$\;
$\Sigmamat \longleftarrow \Sigmamat  - L \times \Deltamat$ \;
}
\KwRet {${{\Mmat}_\lambda}$}\;
}
\SetKwProg{Fn}{Backward}{:}{}
\Fn{\Backward{$x$}}
{
$y \longleftarrow x$\;
\KwRet $y$\;
}
\end{algorithm}
}
\begin{figure}[th]
   \centering
   \vspace{0mm}
 \includegraphics[width=0.85\linewidth]{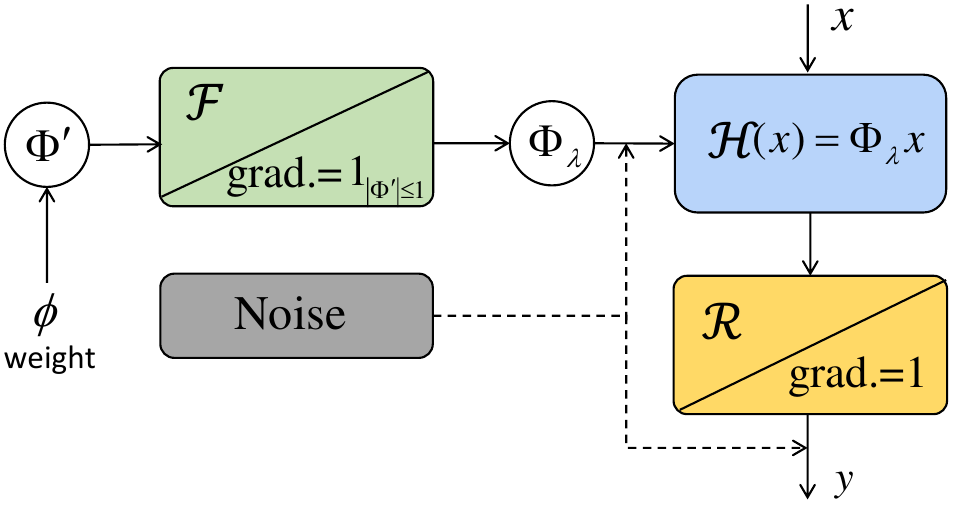}
    \vspace{0mm}
    \caption{\small {Illustration of structural mask optimization. During forward propagation, $\yv = {\cal R}[ {\cal F}({\Phimat}') \cdot \xv]$. During back propagation, the derivative of ${\cal R}$ and ${\cal F}$ are set to 1. Noise should be considered into the encoder when error caused by measurement noise and physical mask miscalibration is non-negligible.}}
    \vspace{0mm}
 \label{fig:encoder}
 \end{figure}

By considering mask as learnable weights, jointly optimizing mask with a deep reconstruction network could contribute to video SCI as demonstrated in previous binary mask optimization works~\cite{ILIADIS2020102591,yoshida2018joint,martel2020neural}, which is generally faced with difficulties in forward-propagation {\em binarization} and back-propagation {\em differentiability}. 
Compared with them, optimizing the proposed structural mask is more challenging due to the difficulties in forward-propagation {\em discretization} and {\em structuralization}, and back-propagation {\em differentiability}.

A pioneering work~\cite{hubara2016binarized} indicated that back-propagation gradients through discretization can be considered to be invariant as long as the forward-propagation input is limited in $[-1, 1]$.
Inspired by this work, we propose a differentiable structural mask transformer ${\cal F}$ to generate ${\lambda}$-bit structural mask ${\Mmat}_{\lambda}$ from a learnable floating-point mask ${\Mmat}' \!\in\! [0,1]$. As illustrated in Alg.~\ref{alg:mask}, the input ${\Mmat}'$ is first discretized and then structuralized in the desired discrete domain during forward propagation, and following the training strategy in~~\cite{hubara2016binarized}, the gradient of ${\cal F}$ is set to $1$ during backward propagation. In the structuralization process, the discretized mask $\Mmat$ is fine-tuned to meet the structure of the temporal sum being $1$ at all spatial position.
The same back propagation strategy is also used for the non-differentiable sensor response ${\cal R}$.
$\Phimat'$ and $\Phimat_{\lambda}$ are the measurement matrix form of $\Mmat'$ and $\Mmat_{\lambda}$, respectively.
The whole mask optimization process is depicted in Fig.~\ref{fig:encoder}.

\subsection{Res2former as Deep Decoder}
\label{sec:decoder}
Building spatial-temporal interactions is the key of video SCI reconstruction. Temporal features are considered to be as important as spatial features in previous works~\cite{Qiao2020_APLP,Wang2021_CVPR_MetaSCI,Cheng2021_CVPR_ReverSCI,wu2021dense,yang2022ensemble,wang2022spatial}.
Previous SOTA STFormer~\cite{wang2022spatial} has a great ability of capturing long-term spatial-temporal interactions but its computational complexity and memory occupation is too high to enable real-world large-scale video SCI.
Considering the motion-aware measurement in Fig.~\ref{fig:mask} (d) caused by the proposed structural mask, we tailor an highly efficient reconstruction network, dubbed Res2former, as the deep decoder. Res2former is the first to put most computations in capturing long-term temporal dependencies.

As demonstrated in the decoder of Fig.~\ref{fig:framework} (b),
Res2former is composed of a feature extraction module, a two-level U-shaped network built by multiple ResTSA modules, and video recovery module.
Feature extraction module is to extract low-level features from measurement domain, composed of two 3D convolutional layers with kernel sizes of $3 \!\times\! 3 \!\times\! 3$ and $1 \!\times\! 3 \!\times\! 3$ respectively.
The video reconstruction module is composed of pixelshuffle~\cite{shi2016real} and two 3D convolution layers with kernel sizes of $1 \!\times\! 1 \!\times\! 1$ and $3 \!\times\! 3 \!\times\! 3$ respectively.
From the perspective of U-net~\cite{ronneberger2015u}, $N_1$ ResTSA modules work with two downsampling/upsampling operations as encoder/decoder and $N_2$ ResTSA modules as the bottleneck.
Such an architecture can enable Res2former to learn high-level feature residuals from the low-level features computationally efficiently.
The main novelties of Res2former is ResTSA module and its temporal self-attention (TSA) mechanism. Next, we introduce them in detail. 

{\noindent \bf ResTSA Module.}
Previous works \cite{xie2017aggregated, ma2018shufflenet} have indicated that channel grouping calculations can effectively reduce model complexity and layered interactions between groups can effectively improve the multiple-scale representation ability \cite{gao2019res2net}. As shown in Fig.~\ref{fig:framework} (c), ResTSA module is also a hierarchical and residual-like structure built by multiple TSA branches. Given an input $\Lmat$, a $P$-level ResTSA module can be formulated as
\begin{equation}
\setlength{\abovedisplayskip}{7pt}
\setlength{\belowdisplayskip}{7pt}
\hspace{-5mm}   \begin{array}{l}
   {\Lmat_{1}, \Lmat_{2}, ..., \Lmat_{P} = {\tt Div}(\Lmat),}\\
   {\Lmat'_{1} = {\rm TSA}(\Lmat_{1}),}\\
   {\Lmat'_{2} = {\rm TSA} (\Lmat_{2} + \Lmat'_{1}), }\\
   \qquad {\vdots}\\
   {\Lmat'_{P} = {\rm TSA} (\Lmat_{P} + \Lmat'_{P-1}),}\\
   {\bar \Lmat} = \Lmat + {\tt Conv}_{1 \!\times\! 1 \!\times\! 1}({\tt Concat}(\Lmat'_{1},\Lmat'_{2},...,\Lmat'_{P})),
   \end{array}
\end{equation}
where $\tt Div$ and $\tt Concat$ denote the channel division and concatenate respectively.

{\noindent \bf TSA Branch.}
With the global perception ability, Transformer can mitigate the
shortcomings caused by CNNs' limited receptive field and has achieved SOTA performance for video SCI reconstruction~\cite{wang2022spatial}. However, self-attention computation along spatial-temporal (3D) dimensions leads to a computational bottleneck for real-world large-scale video SCI applications. Inspired by \cite{bertasius2021space, wang2022spatial}, we limit self-attention mechanism to the temporal dimension for each ResTSA module. Given an input $\Lmat \!\in\! \mathbb{R}^{B \!\times\! H \!\times\! W \!\times\! C}$, a 2D convolution is first used to establish local interactions and then the output is reshaped into $\Lmat_{t} \!\in\! \mathbb{R}^{ HW \!\times\! B \times C}$, \ie, ${ \Lmat_{t}= {\tt Reshape}({\tt Conv}_{3 \!\times\! 3}(\Lmat))}$.
Next, we can obtain \emph{query} $\Qmat \!\in\! \mathbb{R}^{HW \!\times\! B \!\times\! \frac{C}{2}}$, \emph{key} $\Kmat \!\in\! \mathbb{R}^{HW \!\times\! B \!\times\! \frac{C}{2}}$, and \emph{value} $\Vmat \!\in\! \mathbb{R}^{HW \!\times\! B \!\times\! \frac{C}{2}}$ by the following linear projection:
\begin{equation}
\setlength{\abovedisplayskip}{7pt}
\setlength{\belowdisplayskip}{7pt}
\Qmat = \Lmat_t \Wmat^{Q}, \\
\Kmat = \Lmat_t \Wmat^{K},\\
\Vmat = \Lmat_t \Wmat^{V},
\end{equation}
where $\{\Wmat^{Q}, \Wmat^{K}, \Wmat^{V}\} \!\in\! \mathbb{R}^{C \!\times\! \frac{C}{{2}}}$ denote the linear projection matrices. Note that the output dimension is reduced to half of the input dimension, further decreasing the computational complexity.
Then, $\Qmat$, $\Kmat$, and $\Vmat$ are divided into $N$ heads along the feature channel:
$\Qmat \!=\! \{\Qmat_k\}_1^N$, $\Kmat \!=\! \{\Kmat_k\}_1^N$, $\Vmat \!=\! \{\Vmat_k\}_1^N \!\in\! \mathbb{R}^{HW \!\times\! B \!\times\! \frac{C}{2N}}$. For $k$-th \emph{head}, the attention can be calculated by 
\begin{equation}
\setlength{\abovedisplayskip}{7pt}
\setlength{\belowdisplayskip}{7pt}
head_k = \Amat_k * \Vmat_k,
\end{equation}
where $\Amat_{k} \!=\! {\tt softmax}({\Qmat_j}{\Kmat_j^{T}/
{\sqrt d }}) \!\in\! \mathbb{R}^{HW \!\times\! B \!\times\! B}$ represents an attention map with a scaling parameter $d = \frac{C}{2N}$. Finally, we concatenate
the outputs of $N$ \emph{heads} along the channel dimension and perform a linear mapping to obtain the final output ${\Lmat'} \!\in\! \mathbb{R}^{B\!\times\! H \!\times\! W \!\times\! C}$: 
\begin{equation}
\setlength{\abovedisplayskip}{7pt}
\setlength{\belowdisplayskip}{7pt}
\Lmat' \!=\! {\Lmat} \!+\! \mathtt{Reshape}(\Wmat(\mathtt{Concat}[head_1,...,head_N])),
\end{equation}
where $\Wmat \!\in\! \mathbb{R}^{ \frac{C}{2}\!\times\! C}$ is the linear projection matrix.
After temporal self-attention calculations, long-term correlation have been established. Next, we use the feed-forwad network, composed of two 3D convolutions with kernel sizes of  $3 \!\times\! 3 \!\times\! 3$ and $1 \!\times\! 1 \!\times\! 1$, respectively, to further improve the model capacity and the local detail refinement ability, which can be formulated as
\begin{equation}
\setlength{\abovedisplayskip}{7pt}
\setlength{\belowdisplayskip}{5pt}
{\bar \Lmat}= {\Lmat'} \!+\! {\tt Conv}_{1 \!\times\! 1\!\times\! 1}({\tt LeakyReLU} ({\tt Conv}_{3 \!\times\! 3\!\times\! 3}({\Lmat'}))).
\end{equation}
 
\subsection{Compared with Previous Framework}
Previous reconstruction networks~\cite{Qiao2020_APLP,Cheng20ECCV_Birnat,Wang2021_CVPR_MetaSCI,Cheng2021_CVPR_ReverSCI,wang2022spatial,meng2023deep,Ma19ICCV,wu2021dense,yang2022ensemble} were trained using random binary mask without considering sensor response and thus have achieved impressive performance on simulation rather than real system. The proposed deep optics framework aims to remove the gap. As mentioned previously, the modeled encoder is essential for training and simulated testing. Following the definitions of the encoder in Tab.~\ref{tab:define}, our deep optics framework and previous framework have differences in
\begin{itemize}
\setlength{\itemsep}{-1pt}
   \item Previous framework: $i)$ training a deep decoder with the {\it RBw/oSR} encoder; $ii)$ deploying the well-trained deep decoder into real video SCI system.
   \item Our framework: $i)$ training a deep decoder with the {\it LSw/SR} encoder in an E2E fashion; $ii)$ deploying the learned structural mask and the well-trained deep decoder into real video SCI system.
\end{itemize}

\begin{table}
\caption{Definition of different encoders.}
\label{tab:define}
\vspace{1mm}
\resizebox{0.48\textwidth}{!}
{
\begin{tabular}{c|c}
\hline
Encoder & Configuration \\
\hline
{\it RBw/oSR} & {Random Binary Mask without Sensor Response}\\
{\it RBw/SR} & {Random Binary Mask with Sensor Response} \\
{\it LSw/SR} & {Learned Structural Mask with Sensor Response}\\\hline
\end{tabular}
}
\end{table}
\begin{table}
\vspace{-2mm}
\caption{Average PSNR (left), SSIM (center) and Q-Score (right) of different networks on six grayscale benchmark datasets.
}
\vspace{1mm}
\centering
\resizebox{0.485\textwidth}{!}
{
\centering
\begin{tabular}{c|c|c}
\hline
\multirow{2}{*}{{Network}}
& Train: {\it RBw/oSR}
& Train: {\it RBw/oSR} \\
& Test: {\it RBw/oSR}
& Test: {\it RBw/SR} \\
\hline
\hline
E2E-CNN~\cite{Qiao2020_APLP} 
& 29.45, 0.882, 47.31
& 27.02, 0.878, 46.82
\\
\hline 
BIRNAT~\cite{Cheng20ECCV_Birnat} 
& 33.31, 0.951, 50.30
& 29.72, 0.935, 48.80
\\
\hline
MetaSCI~\cite{Wang2021_CVPR_MetaSCI} 
& 31.72, 0.926, 48.34
& 28.84, 0.921, 47.92
\\
\hline
RevSCI~\cite{Cheng2021_CVPR_ReverSCI}
& 33.92, 0.956, 51.21
& 29.71, 0.939, 49.43
\\
\hline
SCI3D~\cite{wu2021dense}
& 35.26, 0.968, 52.70
& 30.97, 0.952, 50.94
\\
\hline
ELP-Unfolding~\cite{yang2022ensemble}
& 35.41, 0.969, 53.02
& 30.77, 0.955, 51.53
\\
\hline
STFormer~\cite{wang2022spatial}
& 36.34, 0.974, 54.00
& 31.78, 0.962, 52.15
\\
\hline 
\end{tabular}
}
\vspace{-3mm}
\label{tab:degrade}
\end{table}
\begin{table}
\caption{Average PSNR, SSIM and Q-Score of different re-trained networks on six grayscale benchmark datasets.}
\vspace{1mm}
\centering
\resizebox{0.3\textwidth}{!}
{
\centering
\begin{tabular}{c|c}
\hline
{Network}
& Train \& Test: {\it RBw/SR}\\
\hline
\hline
E2E-CNN\cite{Qiao2020_APLP} 
& 24.67, 0.878, 46.25
\\
\hline 
RevSCI \cite{Cheng2021_CVPR_ReverSCI}
& 26.46, 0.897, 46.57
\\
\hline
SCI3D \cite{wu2021dense}
& 27.54, 0.939, 49.51
\\
\hline
STFormer \cite{wang2022spatial}
& 27.66, 0.941, 49.69
\\
\hline 
\end{tabular}
}
\vspace{-3mm}
\label{tab:retrain}
\end{table}
\begin{figure}
   \centering
  \vspace{0mm}
\includegraphics[width=0.98\linewidth]{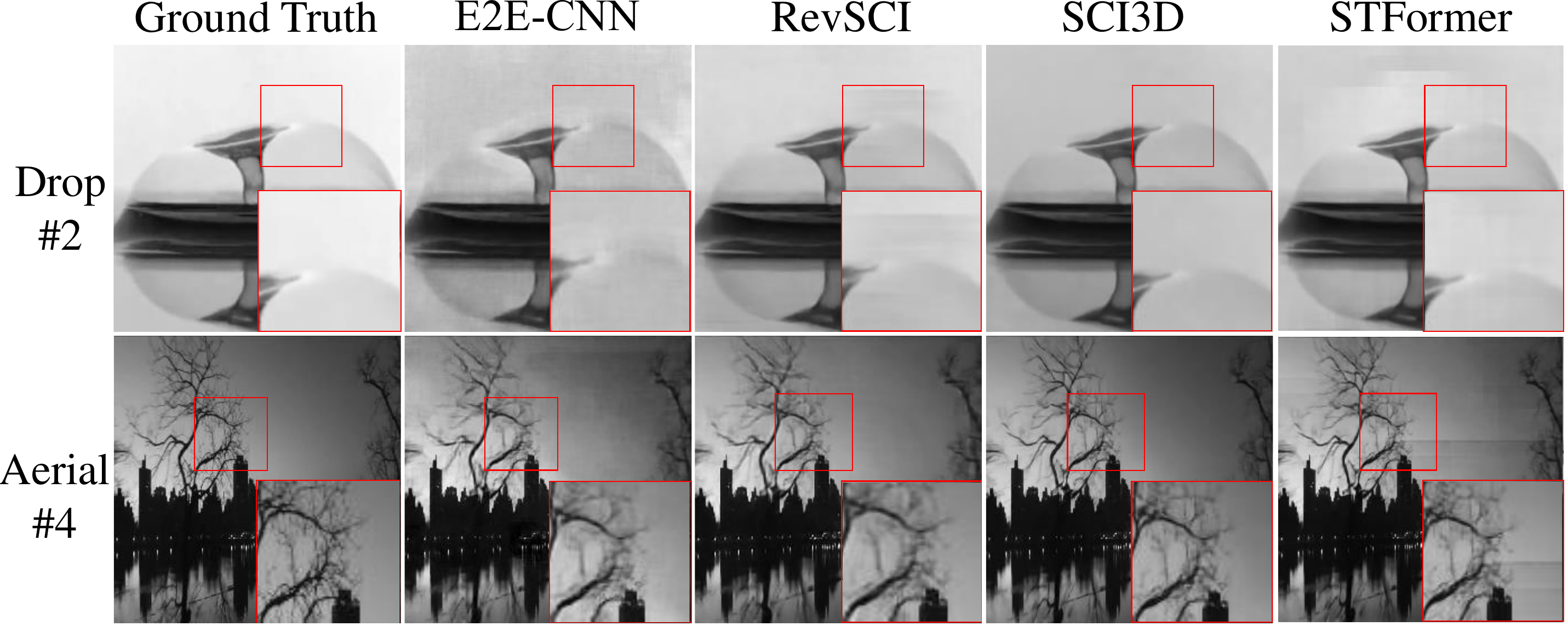}
       \vspace{0mm}
    \caption{Visual results of the re-trained E2E-CNN, RevSCI, SCI3D, and STFormer.}
       \vspace{-5mm}
 \label{fig:retrain}
 \end{figure}
\section{Experiments}
In this section, we validate the effectiveness of the proposed deep optics framework with learnable structural mask and reconstruction network Res2former.
We evaluate the accuracy of different reconstruction networks by peak signal-to-noise-ratio (PSNR), structured similarity index metrics (SSIM), and Q-Score of HDR-VDP-2~\cite{MantiukHDR2011} that measures the dynamic range and by our built {\bf real system}.


\subsection{Datasets and Implementation Details}
Following previous works~\cite{Qiao2020_APLP,Cheng20ECCV_Birnat,Wang2021_CVPR_MetaSCI,Cheng2021_CVPR_ReverSCI,wang2022spatial,meng2023deep,Ma19ICCV,wu2021dense,yang2022ensemble}, we employ DAVIS2017~\cite{pont20172017} as the training dataset.
For the simulation test, 6 benchmark datasets including {\tt Kobe}, {\tt Runner}, {\tt Drop}, {\tt Traffic}, {\tt Aerial}, and {\tt Vehicle} with a size of $256 \!\times\! 256 \!\times\! 8$ are used.
For the real data, we built a video SCI prototype by implementing the learned structural mask on
a DLP7000 DMD, whose details are in supplementary materials (SM).
The real data with a size of $768 \!\times\! 1024 \!\times\! 10$ are captured from two scenes: {\tt Car} and {\tt Windmill}.
We implement the proposed method by PyTorch and all models are trained with Adam optimizer on 8 A40 GPUs.
The initial learning rate is ${1\times10^{-4}}$ and gradually reduced to ${1\times10^{-5}}$.

 \begin{table*}[!htbp]
   \caption{Average PSNR, SSIM, Q-Score, Parameters, FLOPs, and running time of different networks under previous framework or our deep optics framework on six grayscale benchmark datasets.}
   \vspace{1mm}
   \centering
   \resizebox{1\textwidth}{!}
   {
   \centering
   \begin{tabular}{c|c|c|c|c|c|c}
   \hline
   \multirow{2}{*}{{Network}}
   &  Under Previous Framework
   &  Under Our Framework 
   &  \multirow{2}{*}{Gain $\uparrow $} 
   &  Parameters
   &  FLOPs 
   &  Runing Time \\
   &  (with impractical {\it RBw/oSR} encoder)
   &  (with practical {\it LSw/SR} encoder)
   & 
   &  (M)
   &  (G)
   &  (s)\\

   \hline
   \hline
   E2E-CNN~\cite{Qiao2020_APLP} 
   & 29.45, 0.882, 47.31
   & 32.42, 0.940, 49.72
   & 2.97, 0.058, 2.41
   & 0.82
   & 53.49
   & 0.01
   \\
   \hline 
   RevSCI~\cite{Cheng2021_CVPR_ReverSCI}
   & 33.92, 0.956, 51.21
   & 34.81, 0.965, 52.74
   & 0.89, 0.009, 3.31
   & 5.66
   & 766.95
   & 0.19 
   \\
   \hline
 STFormer~\cite{wang2022spatial}
   & 36.34, 0.974, 54.00
   & 36.69, 0.976, 55.08
   & 0.35, 0.002, 1.08
   & 19.48
   & 3060.75
   & 0.49
   \\
   \hline
   \rowcolor{lightgray}Res2former
   & NA
   & 35.98, 0.972, 54.31 
   & NA
   & 11.02
   & 861.76
   & 0.19
   \\
   \hline
   Res2former-L
   & NA
   & 36.56, 0.975, 54.93
   & NA
   & 17.70
   & 1362.51
   & 0.42
   \\
   \hline
   \end{tabular}
   }
   \vspace{0mm}
   \label{tab:summary}
 \end{table*}

\begin{figure*}[th]
   \centering
    \includegraphics[width=1\linewidth]{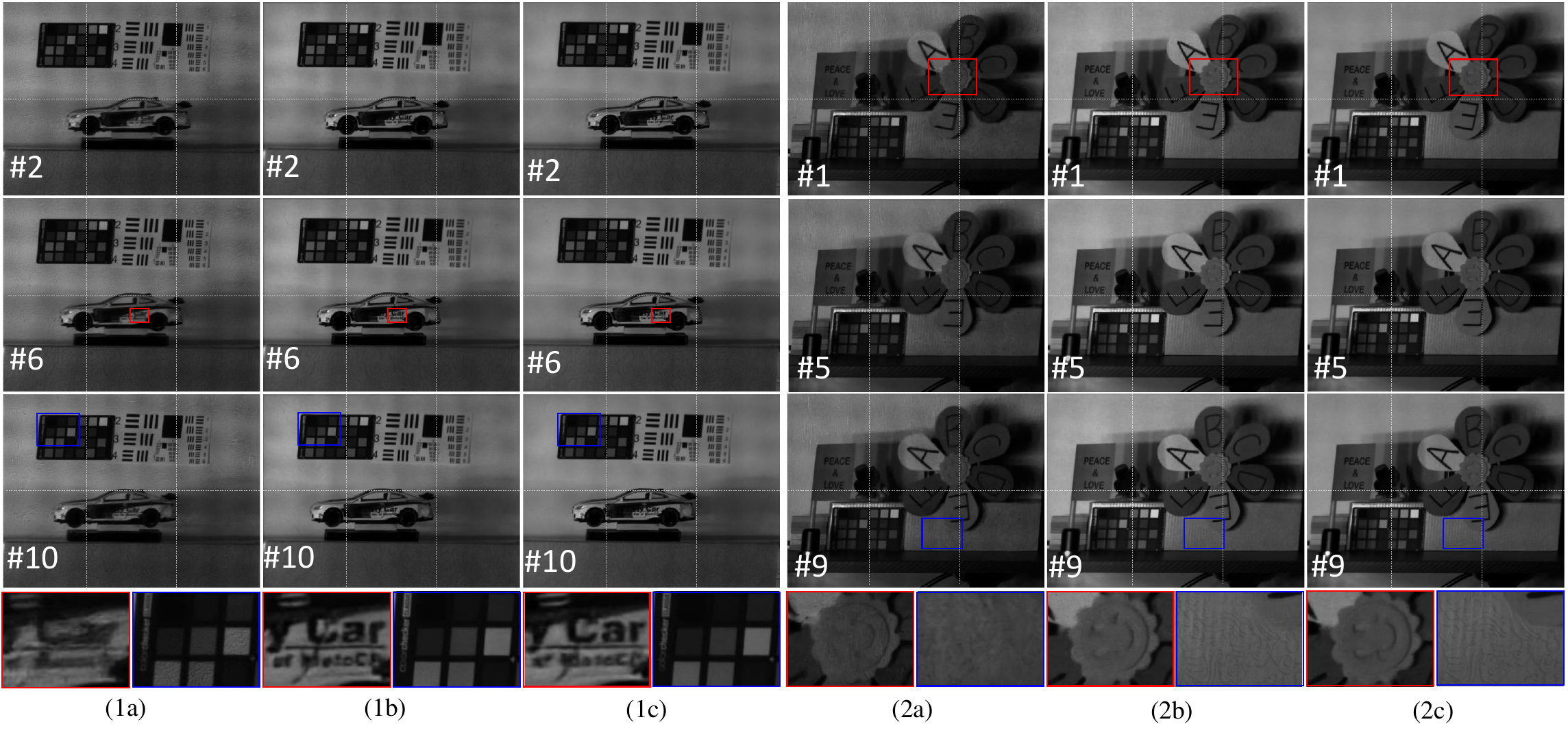}
    \vspace{-5mm}
    \caption{Real reconstructed results using our built video SCI system. (1a)-(2a), (1b)-(2b), and (1c)-(2c) are reconstructed by STFormer under previous framework, STFormer under our framework, and Res2former under our framework, respectively. Our deep optics framework brings a significant improvement compared with previous framework. In real data, the proposed Res2former is as good as STFormer.}
    \vspace{-3mm}
 \label{fig:real_result}
 \end{figure*}
\begin{figure}[th]
   \centering
\includegraphics[width=1\linewidth]{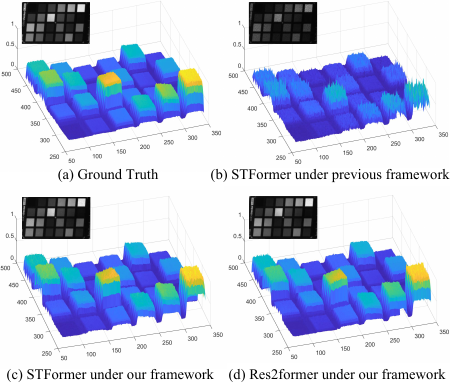}
    \vspace{0mm}
    \caption{Dynamic range comparison through the 3D heat map of a standard ColorChecker placed in scenes, which is an average effect of the recovered $10$ video frames. Obviously, Res2former and STFormer under our framework can retrieve wider dynamic range (close to ground truth) than STFormer under previous framework.}
    \vspace{-4mm}
 \label{fig:hdr}
\end{figure}
\subsection{Results on Synthetic Data}
Before evaluating the proposed method, we give a clear insight into previous networks' performance degradation in real system, including E2E-CNN~\cite{Qiao2020_APLP}, BIRNAT~\cite{Cheng20ECCV_Birnat}, MetaSCI~\cite{Wang2021_CVPR_MetaSCI}, RevSCI~\cite{Cheng2021_CVPR_ReverSCI}, SCI3D~\cite{wu2021dense}, ELP-Unfolding~\cite{yang2022ensemble}, and previous SOTA STFormer~\cite{wang2022spatial}. Their pre-trained models were originally trained on the {\it RBw/oSR} encoder and now are tested with the {\it RBw/SR} encoder closer to real system. To avoid overexposure caused by binary mask, automatic aperture is simulated by scaling before sensor response. 
As shown in Tab.~\ref{tab:degrade}, there is a serious degradation in both structural information (PSNR, SSIM) and dynamic range (Q-Score). Obviously, it is caused by the mismatch between training without sensor response and testing with sensor response. Unfortunately, the {\it RBw/oSR}-training-plus-{\it RBw/SR}-using framework is prevailing even though sensor response is inevitable in hardware encoder.
It could be a natural explanation for why the performance of previous networks~\cite{Qiao2020_APLP,Cheng20ECCV_Birnat,Wang2021_CVPR_MetaSCI,Cheng2021_CVPR_ReverSCI,wang2022spatial,meng2023deep,Ma19ICCV,wu2021dense,yang2022ensemble} on real data is greatly inferior to that on simulation. Moreover, we have re-trained four representative networks, including E2E-CNN~\cite{Qiao2020_APLP}, RevSCI~\cite{Cheng2021_CVPR_ReverSCI}, SCI3D~\cite{wu2021dense}, and STFormer~\cite{wang2022spatial} with the {\it RBw/SR} encoder. As shown in Tab.~\ref{tab:retrain} and Fig.~\ref{fig:retrain}, the re-trained networks lead to worse results and their reconstructed results have an clear degradation in dynamic range. It is because these networks are forced to resolve a hybrid problem of compressed video reconstruction and dynamic range reconstruction. With the limited dynamic range of image sensor in real system, random binary mask is therefore not the best choice. Using high-bit-depth image sensor could alleviate this problem but its high cost clashes with the virtues of video SCI. 

Next, we validate the generalization of our deep optics framework and the effectiveness of Res2former on balancing reconstruction performance and computational load. E2E-CNN~\cite{Qiao2020_APLP}, RevSCI~\cite{Cheng2021_CVPR_ReverSCI}, and STFormer~\cite{wang2022spatial} are re-trained under our framework to replace Res2former and $4$-bit structural mask is jointly optimized by default. As shown in Tab.~\ref{tab:summary}, three other networks have archived improvement with various degrees. Under our framework, E2E-CNN has achieved a significant improvement in PSNR and SSIM and RevSCI has achieved a best improvement in terms of Q-Score. Compared with STFormer under our framework, Res2former can achieve the competitive performance ($\!<\!1$ dB) with only $28.15\%$ FLOPs and $56.57\%$ parameters of STFormer and far less running time than STFormer. 
We have also tried to increase the parameters of Res2former to the level of STFormer. The large Res2former, dubbed Res2former-L, is got by increasing channels from $96$ to $128$ and the depth of ResTSA module from $N_1\!=\!3$ to $N_1\!=\!5$. Res2former-L can achieve the same-level performance with STFormer under our framework and outperforms the original STformer.

\subsection{Results on Real Data}
We validate the effectiveness of the proposed deep optics framework and Res2former in our built video SCI system whose details are in SM.
Previous SOTA STFormer is regarded as the benchmark reconstruction network.
We conduct real system test in the following three settings: $i)$ STFormer under previous framework; $ii)$ STFormer under our framework; $ii)$ Res2former under our framework. Two kinds of high-speed scenes ({\tt Car} and {\tt Windmill}) are modulated by random binary masks or the learned structural masks and then integrated into single-shot $768\!\times\! 1024$ measurement frames by an off-the-shelf image sensor with $50$ fps. The compressed frames is $10$. To ensure motion uniformity, {\tt Car} and {\tt Windmill} are driven by an electric linear gateway and a rotating motor, respectively. As shown in Fig.~\ref{fig:real_result}, 
the reconstructed results of STFormer under our framework is far better than that of the original STFormer in terms of dynamic and static regions. Under our framework, the results of STFormer and Res2former are visually close. However, STFormer’s GPU memory occupation and inference time are $33$ GB and $4.55$ s but only $16$ GB and $2.10$ s for Res2former. Moreover, we analyze the dynamic range of the reconstructed results. As demonstrated in Fig.~\ref{fig:hdr}, the proposed framework can eliminate the dynamic range degradation rooted in previous works completely and achieve FDR video SCI. More results are in SM.

\subsection{Ablation Study}
\begin{table}[hbt]
   \vspace{-2.5mm}
   \caption{Ablation study with different kinds of structural mask.}
   \vspace{1mm}
   \centering
   \resizebox{0.368\textwidth}{!}
   {
   \centering
   \begin{tabular}{c|c|c}
   \hline
   {Mask}
   & Random
   & Learned \\
   \hline
   $1$-bit
   & 34.25, 0.964, 50.06
   & 34.62, 0.964, 51.45
   \\
    \hline
   $2$-bit
   & 35.03, 0.967, 53.17
   & 35.87, 0.971, 53.91
   \\
   \hline 
   $3$-bit
   & 34.91, 0.965, 52.54
   & 35.90, 0.971, 54.17
   \\
   \hline
   $4$-bit
   & 34.95, 0.965, 52.85
   & 35.98, 0.972, 54.31
   \\
   \hline
   \end{tabular}
   }
\label{tab:ablation}
\vspace{-3mm}
\end{table}
To offer an insight into structural mask optimization, we conduct two ablation experiments on our deep optics framework: $i)$ with different-bit structural mask; $ii)$ with learnable or random structural mask. All experiments are tested on six grayscale benchmark datasets. As shown in Tab.~\ref{tab:ablation}, mask optimization can contribute to reconstruction regardless of the bit depth of structural mask. The larger the learnable mask bit depth is, the better the performance is. This conclusion is consistent with mask conditioning in~\cite{vargas2021time}. With randomly generated structural mask, Res2former cannot achieve its full potential.

\section{Conclusion}
Aiming to move one step further of video SCI towards practical applications, we have proposed a deep optics framework to jointly optimize the proposed structural mask and reconstruction network Res2former. As validated in simulation and real system, our framework can bring a significant improvement for other networks. Besides, our Res2former can provide competitive performance in a computationally efficient manner.

\vspace{1.5mm}

\noindent \textbf{Acknowledgements:} This work was supported by National Natural Science Foundation of China (62271414), Zhejiang Provincial Natural Science Foundation of China (LR23F010001) and Research Center for Industries of the Future (RCIF) at Westlake University.

{\small
\bibliographystyle{ieee_fullname}
\bibliography{egbib}
}

\end{document}